%% file: root.tex
\documentclass[letterpaper, 10 pt, conference]{ieeeconf}  

\IEEEoverridecommandlockouts                              

\usepackage{amsmath} 
\usepackage{amssymb}  
\usepackage{graphicx}
\usepackage{caption}
\usepackage{cuted}      
\usepackage{booktabs}
\usepackage{float}
\usepackage{xcolor}
\usepackage{hyperref}
\title{\textbf{PhyGile}: Physics-Prefix Guided Motion Generation \\for Agile General Humanoid Motion Tracking
}

\author{Jiacheng Bao$^{1,2*}$, Haoran Yang$^{2,3*}$, Yucheng Xin$^{2,4*}$, Junhong Liu$^{1}$, Yuecheng Xu$^{5}$, Han Liang$^{6}$\\
Pengfei Han$^{2}$, Xiaoguang Ma$^{7}$, Dong Wang$^{2}$, Bin Zhao$^{1,2\dagger}$
\thanks{$^{1}$Northwestern Polytechnical University, $^{2}$Shanghai AI Laboratory, $^{3}$University of Science and Technology of China, $^{4}$Tsinghua University, $^{5}$Fudan University, $^{6}$ByteDance, $^{7}$Northeastern University}
\thanks{$^{*}$indicates equal contribution, $^\dagger$Corresponding author}
}

\begin{document}

\maketitle
\thispagestyle{empty}
\pagestyle{empty}

\input{sections/0abstract}

\input{sections/1intro}
\input{sections/2related}
\input{sections/3methods}

\input{sections/4exp}
\input{sections/5conclusion}

\bibliographystyle{IEEEtran}
\bibliography{references}

\input{sections/appendix}

\end{document}

%% file: sections/0abstract.tex
\begin{strip}
\vspace{-50pt}
\centering
\includegraphics[width=\linewidth]{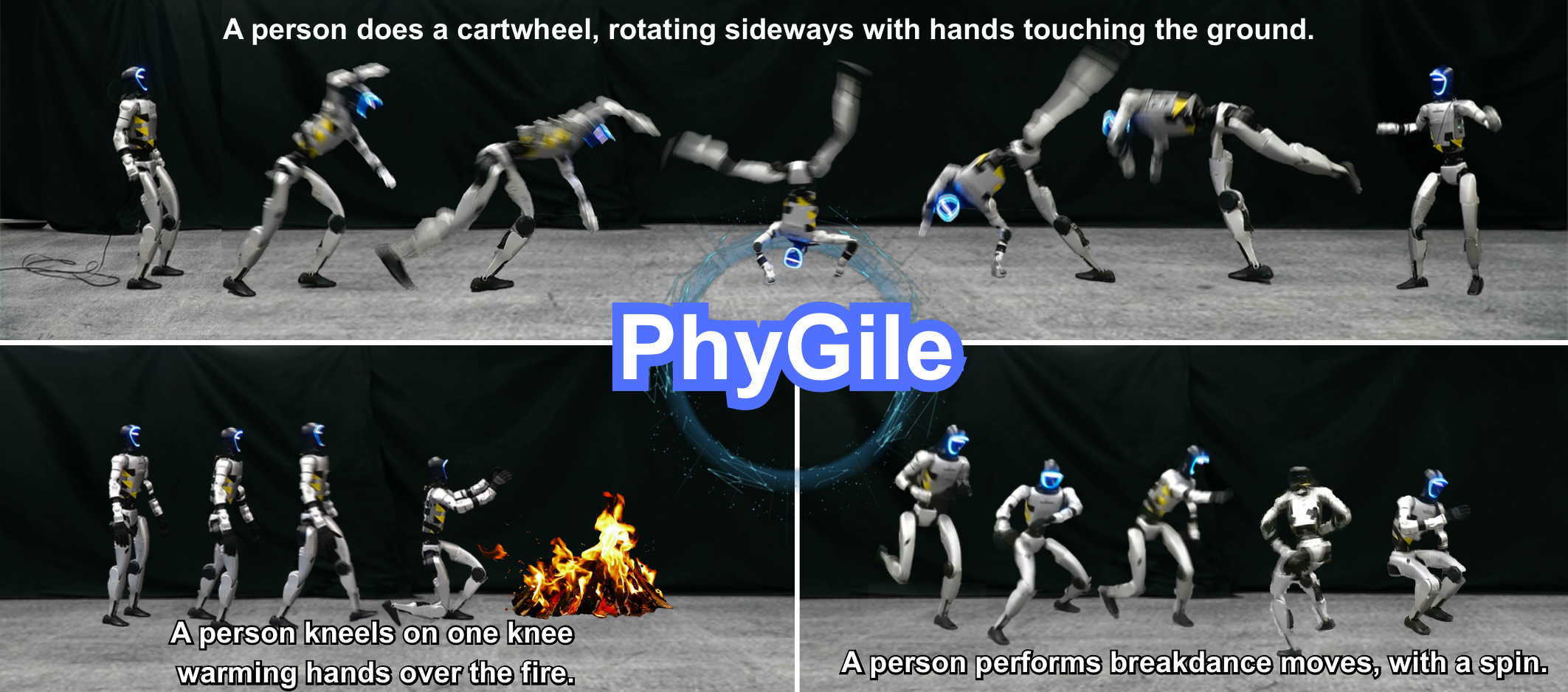}
\captionof{figure}{\textbf{PhyGile} translates natural language commands into agile and expressive whole-body motions on humanoid robots, thereby enabling stable real-world execution of highly-difficult motions. \textbf{Project Page:} \href{https://baojch.github.io/phygile-page/}{\textcolor{blue!65}{\textbf{baojch.github.io/phygile-page/}}}}
\label{fig:teaser}
\end{strip}

\begin{abstract}
    Humanoid robots are expected to execute agile and expressive whole-body motions in real-world settings. Existing text-to-motion generation models are predominantly trained on captured human motion datasets, whose priors assume human biomechanics, actuation, mass distribution, and contact strategies. When such motions are directly retargeted to humanoid robots, the resulting trajectories may satisfy geometric constraints (e.g., joint limits, pose continuity) and appear kinematically reasonable. However, they frequently violate the physical feasibility required for real-world execution.
    To address these issues, we present PhyGile, a unified framework that closes the loop between robot-native motion generation and General Motion Tracking (GMT).
    PhyGile performs Physics-prefix-Guided robot-native motion generation at inference time, directly generating robot-native motions in a 262-dimensional skeletal space with physics-guided prefix, thereby eliminating inference-time retargeting artifacts and reducing generation–execution discrepancies.
    Before physics-prefix adaptation, we train the GMT controller with a curriculum-based mixture-of-experts scheme, followed by post-training on unlabeled motion data, to improve robustness over large-scale robot motions.
    During physics-prefix adaptation, the GMT controller is further fine-tuned with generated objectives under physics-derived prefixes, enabling agile and stable execution of complex motions on real robots.
    Extensive offline and real-robot experiments demonstrate that our PhyGile expands the frontier of text-driven humanoid control, enabling stable tracking of agile, highly-difficult whole-body motions that go well beyond walking and low-dynamic motions typically achieved by prior methods.
\end{abstract}

%% file: sections/1intro.tex
\section{Introduction}
\label{sec:intro}
Humanoid control is undergoing a paradigm shift from task-specific locomotion toward scalable and general-purpose motion generation. Recent advances in general motion tracking (GMT) show that a single policy can imitate large collections of reference motions and reliably transfer them to real hardware~\cite {chen2025general,yang2025efficiently,wang2026general}. In parallel, text-driven motion generation has progressed rapidly, with diffusion-based models producing semantically rich and diverse human motions conditioned on natural language~\cite{tevet2022human,zhang2022motiondiffuse,chen2023executing}.
Recently, these two paradigms have begun to converge, where text-to-motion models provide high-level motion priors and GMT policies serve as low-level executors, forming a hierarchical pipeline for autonomous humanoid control~\cite{xie2026textop,jiang2025uniact,shao2025langwbc}.

Despite the above-mentioned successes, motion generation and physical execution remain fundamentally misaligned in practice. Most text-to-motion models are developed in the human domain and synthesize motions in standardized representations such as SMPL~\cite{SMPL:2015}. Running these motions on humanoid robots requires a retargeting stage~\cite{joao2025gmr, Luo2023PerpetualHC} that maps human joint trajectories to robot morphologies. While forward-kinematics-based retargeting can approximately preserve pose structure, it does not enforce coupled physical constraints, including torque limits, contact consistency, and dynamic balance. Consequently, retargeted motions often contain physically inconsistent segments that appear kinematically plausible yet are dynamically unstable~\cite{peng2018deepmimic}. In large-scale training pipelines, such references degrade policy learning and often necessitate filtering mechanisms to discard invalid samples, reducing data efficiency and robustness.

Beyond retargeting artifacts, GMT is further constrained by severe data imbalance. Large motion datasets exhibit pronounced long-tail distributions. For example, HumanML3D~\cite{Guo_2022_HUMANML3D}, one of the most widely used benchmarks for text-driven motion modeling, contains abundant simple motions such as walking and running, whereas complex and agile motions are comparatively rare. When training a general tracking policy on such skewed distributions, optimization naturally prioritizes frequent and low-difficulty motions. Rare and high-difficulty skills that require tight coordination and dynamic stability remain undertrained, leading to fragile performance when robots attempt agile motions.

Recent humanoid motion generation approaches attempt to reduce reliance on human-first retargeting by aligning language with robot-controllable embeddings~\cite{jiang2025uniact,wang2025sentinel} or directly modeling robot-native motion data~\cite{xie2026textop}. While these methods strengthen the connection between semantics and embodiment, generation and tracking are typically optimized as loosely coupled modules: a generator proposes trajectories, and a tracker attempts to follow them. When generated motions exceed the feasible region of the current controller, execution usually fails unless a mechanism is available to reconcile generation with control.

In this work, we propose PhyGile, a physics-prefix-guided framework that couples robot-native motion generation with agile GMT, enabling high-performance tracking and generation within a single, physically grounded pipeline. With the motion generator frozen, PhyGile leverages physics-validated prefixes as a shared interface to improve motion quality and GMT agility, leading to more physically consistent and task-effective motions at execution time.

On the generation side, we introduce a diffusion-based motion model operating in a 262D robot-skeleton space, directly synthesizing robot-native joint trajectories. To enhance compositional language grounding, we incorporate a Token-level Parameter-mixing Mixture-of-Experts (TP-MoE), enabling fine-grained alignment between individual text tokens and the motion timeline, so that temporally localized semantic cues correspond to appropriate motion segments.

On the control side, we address the long-tail challenge of agile motions through a curriculum mixture-of-experts training scheme. In the first stage, we perform explicit curriculum learning on difficulty-annotated motion data, where motions are stratified by complexity and experts specialize across skill levels. Error-aware adaptive sampling progressively increases exposure to high-difficulty motions, improving robustness on rare and highly dynamic skills. In the second stage, we conduct global soft-moe post-training on large-scale unlabeled motions to enhance overall generalization while preserving the agility acquired during curriculum learning.

To bridge generation and execution, we further introduce a physics-prefix-guided fine-tuning stage. Dynamically feasible motion segments extracted from the tracking policy are injected as conditioning prefixes into the diffusion process, anchoring the initial denoising states to the robot’s motion manifold. During this stage, the diffusion model remains frozen, while the GMT controller is fine-tuned under prefix-conditioned motion generation to improve tracking performance on generated motion distributions, thereby tightening the coupling between generation and execution.

By integrating difficulty-aware tracking, robot-native diffusion generation, and physics-prefix-guided alignment into a unified pipeline, \textbf{PhyGile} enables the generation and execution of agile humanoid motions that are both semantically aligned and physically realizable. Extensive offline and real-robot experiments demonstrate that \textbf{PhyGile} effectively resolves the semantics--physics trade-off in generation, establishing a highly robust and unified tracking policy. Notably, for highly dynamic agile motions where prior pipelines fail, our framework rapidly unlocks stable hardware execution through lightweight fine-tuning, demonstrating strong scalability toward complex whole-body skills.

In summary, \textbf{PhyGile}'s main contributions are as follows:
\begin{itemize}
\setlength\itemsep{0em}

\item We propose \textbf{PhyGile}, a physics-prefix-guided framework for general humanoid motion tracking that couples robot-native diffusion-based motion generation with GMT. By closing the loop via physics-consistent prefixes, PhyGile mitigates the mismatch between text-driven motion synthesis and real-world control, enabling agile and dynamically feasible humanoid motions.

\item We develop three key components: (i) a curriculum-based MOE training strategy for agile GMT with motion-difficulty stratification; (ii) a 262D robot-skeleton representation with TP-MoE-enhanced motion generation for fine-grained text conditioning; and (iii) a physics-prefix-guided fine-tuning stage that aligns generated motions with executable tracking policies.

\item Through extensive offline benchmarks and real-robot experiments, we demonstrate that PhyGile effectively resolves the semantics--physics trade-off in generation. Furthermore, our robust, unified tracking policy achieves stable hardware execution of highly dynamic humanoid motions, successfully unlocking complex agile behaviors that prior pipelines fail to realize even with additional tuning.

\end{itemize}

%% file: sections/2related.tex
\section{Related Works}
\label{sec: related}

\subsection{Text-driven Human Motion Generation}
Text-driven motion generation aims to learn the conditional distribution of human motion given natural-language descriptions. Early approaches adopt VAE-style formulations or autoregressive modeling with discrete motion representations~\cite{tevet2022motionclip, zhang2023generating, jiang2023motiongpt, zhang2023t2mgpt}. Recent diffusion-based methods~\cite{tevet2022human, zhang2022motiondiffuse, chen2023executing,liang2024towards} reframe text-to-motion as conditional denoising, improving generative quality and controllability. Follow-up works further enhance efficiency and scalability via masked modeling and latent-consistency acceleration~\cite{dai2024motionlcm}, as well as improve diversity and long-horizon synthesis through retrieval or compositional generation~\cite{zhang2023remodiffuse, shafir2023human}. Complementary directions incorporate cross-modal alignment or physical constraints to strengthen semantic faithfulness and realism~\cite{yuan2023physdiff, tevet2024closd, hong2022avatarclip}.

\subsection{Humanoid Motion Generation and Control}
Humanoid motion generation and control can be broadly categorized into motion synthesis, robot-native imitation learning, and vision-language-action (VLA) policies. A common paradigm generates human motions conditioned on language and retargets them to humanoid morphologies, decoupling semantics from embodiment~\cite{ding2025humanoid, liu2025commanding}. To improve scalability, recent methods align language with structured latent or discrete action spaces, enabling retargeting-free or tokenized generation~\cite{li2025from, li2025have, luo2025sonic}.
In parallel, robot-native imitation learning and VLA approaches directly learn executable policies from motion, demonstration, or egocentric visual inputs, reducing reliance on explicit motion synthesis and bridging perception, reasoning, and control~\cite{xie2026textop, jiang2025uniact, wang2025sentinel, song2025hume, yang2026zerowbc}. However, such approaches still face challenges in motor control and long-horizon consistency.

General motion tracking for humanoid control casts reference following as scalable motion imitation, producing unified policies that cover diverse behaviors and transfer to hardware~\cite{chen2025general, yang2025efficiently, wang2026general, luo2025sonic, sun2026mosaic,ben2025homie}. To mitigate interference across heterogeneous motions, prior work leverages mixture-of-experts, expert partitioning with consolidation, or multi-behavior distillation~\cite{chen2025general, yang2025efficiently, wang2025from, li2026telegate, zhao2025towards}. Shared latent representations over motion, goals, and rewards further support promptable reuse and planning~\cite{li2025zero, liu2025unleashing}, and these trackers can serve as low-level backbones for language-conditioned control~\cite{xie2026textop, luo2025sonic}. However, highly dynamic and agile behaviors remain challenging due to rapid coordination and contact transitions that amplify tracking instability.

%% file: sections/3methods.tex
\section{Methods}
\begin{figure*}[htbp] 
	\centering  
	\includegraphics[width=1\linewidth]{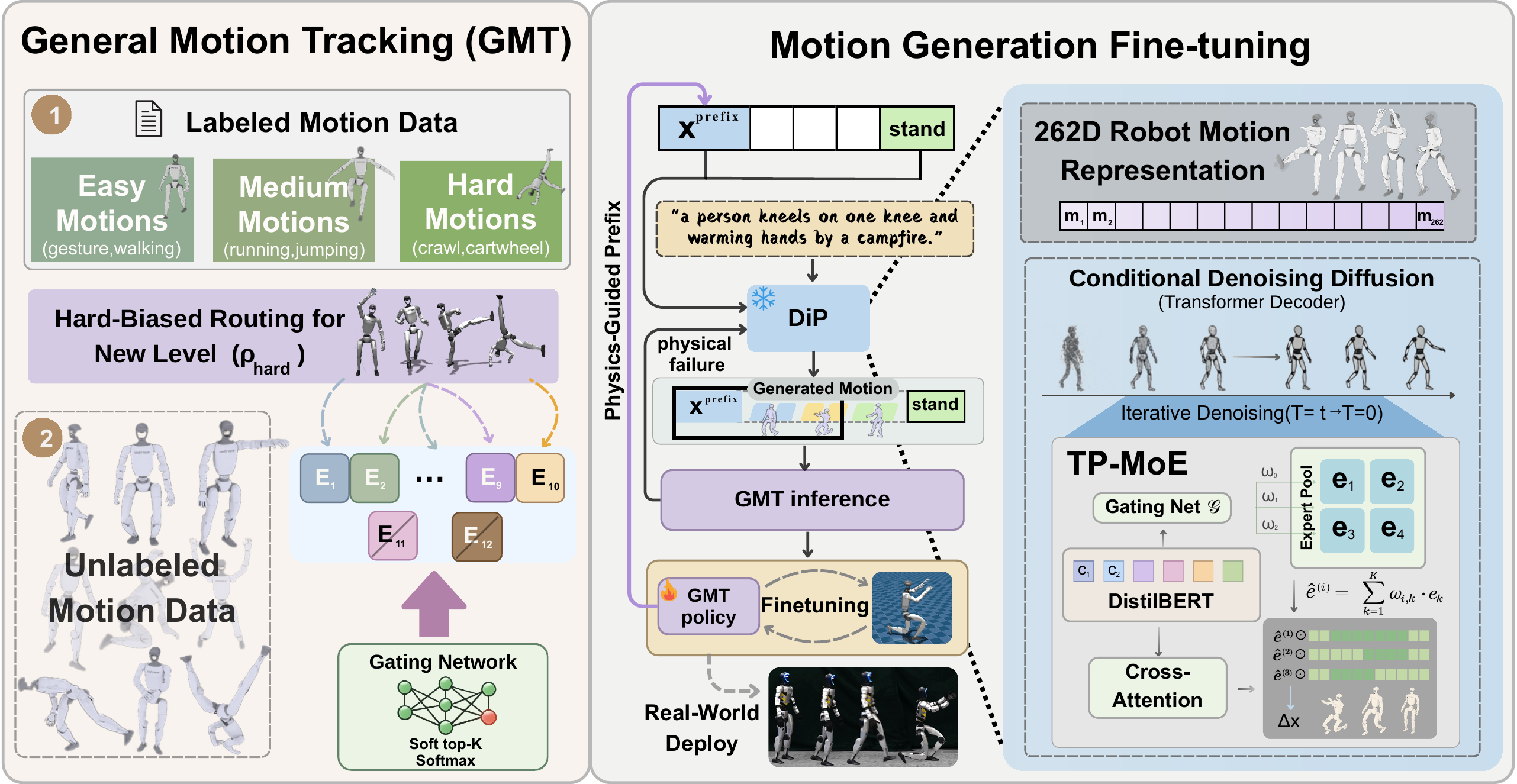} 
	\caption{\textbf{Overview of PhyGile.} \emph{(Left) GMT:} A two-stage MoE tracker is first trained with curriculum-constrained routing to induce expert specialization, followed by global soft post-training with dynamic expert expansion to absorb persistently difficult motions. \emph{(Right) Generation of Diffusion Policy:} A TP-MoE–conditioned robot-native diffusion model generating 262D robot motion sequences from text. \emph{(Center) Motion Generation Fine-tuning:} Executable motion prefixes are concatenated with newly generated 1-second continuations and validated by pretrained GMT. Closed-loop simulation refinement further enforces dynamic feasibility and improves consistency between generated and trackable motions, and the fine-tuned GMT policy is deployed on real robots.
 } 
	\label{fig: method} 
\end{figure*}
\textbf{PhyGile} couples robot-native diffusion generation with agile GMT through physics-prefix guidance. It comprises (i) a two-stage MoE tracker trained with curriculum-constrained routing to cover long-tail agile motions, (ii) a TP-MoE–conditioned robot-native diffusion generator, and (iii) physics-prefix guided adaptation that validates and refines generated segments for dynamic feasibility, tightening consistency between generation and executable tracking.
\subsection{General Motion Tracking}
\label{GMT}
Our training datasets contain approximately 45 hours of motion, including text-annotated sequences from HumanML3D~\cite{Guo_2022_HUMANML3D} and unlabeled MoCap data from AMASS~\cite{mahmood2019amass}, LaFAN1~\cite{harvey2020robust}, plus a private 3-hour MoCap set retargeted to our robot embodiment using GMR~\cite{joao2025gmr}.

\textbf{Data Curriculum.}
We interpret motion difficulty from a neuro-control view as the \emph{coordination load} required for planning, prediction, and stabilization (i.e., increasing coupling and dynamical instability).
Using LLM-based semantic analysis of HumanML3D text descriptions, we assign each clip to one of 12 ordered levels.
Levels 1--10 are feasible in our embodiment and are grouped only for exposition as \emph{Easy} (1--4), \emph{Medium} (5--7), and \emph{Hard} (8--10), while Levels 11--12 are excluded due to environment/physics mismatch.

\noindent\textbf{Level 1: [Easy]} Near-static poses with minimal planning and balance demands (e.g., standing, clapping).

\noindent\textbf{Levels 2--4: [Easy]} Automatic locomotion and basic motor patterns with stable CoM dynamics (e.g., walking, bending).

\noindent\textbf{Levels 5--7: [Medium]} Coordinated transitions and directional or speed changes requiring predictive stabilization (e.g., jogging, walking backward, kicking).

\noindent\textbf{Levels 8--10: [Hard]} Highly dynamic or composite skills with tightly coupled planning and balance (e.g., rolling, jumping, spinning, crawling).

\noindent\textbf{Level 11: [Incompatible]} Motions dependent on external terrain or elevation changes (e.g., stair climbing).

\noindent\textbf{Level 12: [Infeasible]} Motions physically unrealizable in the current simulation setup (e.g., swimming, flying).

\textbf{Freeze-and-drop self-purification.}
During training, for each motion file $i$, we maintain an EMA tracking error $E_i$ and success-rate estimate $\hat{p}_i^{\text{succ}}$ (fraction of rollouts whose tracking succeeds).
A motion is temporarily frozen once it has been sufficiently exposed, but remains untrackable:
\begin{equation}
\left( E_i \geq \tau_{\text{err}} \;\lor\; \hat{p}_i^{\text{succ}} \leq \tau_{\text{succ}} \right) \;\land\; n_i \geq n_{\min},
\end{equation}
where $n_i$ is the number of rollouts sampled from file $i$, $\tau_{\text{err}}$ and $\tau_{\text{succ}}$ are error/success thresholds, and $n_{\min}$ is the minimum exposure before freezing is allowed.
A file is dropped if it repeatedly triggers freezing.

\textbf{Look-ahead motion encoding.}
The policy conditions on multi-scale future motion: a short look-ahead window provides per-frame future targets for accurate near-term tracking, while a longer downsampled horizon is encoded into a compact latent $\mathbf{z}_t$ via temporal convolution and pooling.
$\mathbf{z}_t$ captures upcoming speed changes, turns, and contact events, and is also used as the \emph{only} input to the MoE gate.

\textbf{Two-stage training with MoE routing.}
Our actor is a MoE policy with $K$ expert MLPs $\{E_1,\dots,E_K\}$ and a lightweight gate $G$ that maps $\mathbf{z}_t$ to routing logits $\boldsymbol{\ell}=G(\mathbf{z}_t)\in\mathbb{R}^K$.
To balance routing stability and compute, we use low-frequency soft top-$k$ routing: every $M$ steps we refresh a candidate expert set $\mathcal{K}=\mathrm{top}\text{-}k(\boldsymbol{\ell})$ (thus $|\mathcal{K}|=k$), and between refreshes compute temperature-$\tau$ softmax weights only over candidates.
The action is a convex mixture of candidate expert outputs,
\begin{equation}
\mathbf{a}_t=\sum_{j\in\mathcal{K}} p_j\,E_j(\tilde{\mathbf{o}}_t),
\end{equation}
where $\tilde{\mathbf{o}}_t$ denotes the observation vector and $p_j=\mathrm{softmax}(\boldsymbol{\ell}_{\mathcal{K}}/\tau)_j$ are normalized mixture weights computed from candidate logits $\boldsymbol{\ell}_{\mathcal{K}}$; this requires only $k$ expert forward passes per step.
We optionally apply EMA smoothing to $\boldsymbol{\ell}$ to reduce routing jitter.

\paragraph{Stage I: level-wise curriculum with hard-biased routing}
We train \emph{level-by-level} from $l_{\max}=1$ to $10$, progressively unlocking higher-difficulty motion levels to induce expert specialization.
At the file level, we use difficulty-aware in-batch sampling with uniform exploration, while gradually introducing newly unlocked levels and maintaining a minimum replay quota for earlier levels.

We enforce \emph{hard-biased routing} so the newly unlocked expert receives strong gradients.
When the curriculum is at level $l_{\max}$, routing is restricted to $\{E_1,\dots,E_{l_{\max}}\}$ by masking locked experts.
Moreover, for samples drawn from the current hardest level ($l_i=l_{\max}$, where $l_i$ is the assigned level of file $i$), we bypass the gate with probability $\rho_{\text{hard}}=0.8$ and hard-route to the expert $E_{l_{\max}}$:
\begin{equation}
\mathbf{a}_t=\begin{cases}
E_{l_{\max}}(\tilde{\mathbf{o}}_t), & l_i=l_{\max}\ \land\ u<\rho_{\text{hard}},\\
\sum_{j\in\mathcal{K}} p_j\,E_j(\tilde{\mathbf{o}}_t), & \text{otherwise},
\end{cases}
\end{equation}
where $u\sim\mathrm{Uniform}(0,1)$ and $\rho_{\text{hard}}$ is the hard-routing probability.
Upon level promotion, the new expert is initialized by copying the preceding expert ($\theta_{E_{l}}\!\leftarrow\!\theta_{E_{l-1}}$) to stabilize optimization.
To align motion semantics with routing, we jointly optimize the gate with an auxiliary cross-entropy loss:
\begin{equation}
\mathcal{L}_{\text{route}}=\lambda_{\text{CE}}\cdot \mathrm{CE}\!\left(G(\mathbf{z}_t),\,l_i-1\right),
\end{equation}
where $\mathrm{CE}(\cdot,\cdot)$ is the cross-entropy loss, $(l_i-1)$ is the level-to-expert label, and $\lambda_{\text{CE}}$ is its weight.

\paragraph{Stage II: global soft post-training}
We remove curriculum masks and hard-routing constraints, allowing all $K$ experts to access the full dataset and enabling end-to-end optimization with differentiable soft top-$k$ routing.
To avoid routing collapse, we replace strong level supervision with a load-balancing objective:
\begin{equation}
\mathcal{L}_{\text{bal}}=K\sum_{j=1}^{K} f_j\,\bar{p}_j,
\end{equation}
where $f_j=\mathbb{E}[\mathbb{I}(\arg\max_k p_k=j)]$ is the fraction of samples whose top-1 route selects expert $j$ and $\bar{p}_j=\mathbb{E}[p_j]$ is the mean probability mass assigned to expert $j$ (both estimated over minibatches), encouraging uniform utilization; a low-weight cross-entropy regularizer can be retained to preserve the semantic prior.
To handle distribution shifts in unlabeled data, we additionally support \emph{dynamic expert addition}: we track per-file EMA statistics including routing entropy $\mathcal{H}(p)=-\sum_j p_j\log p_j$ and the top-1/top-2 gap $\Delta=p_{(1)}-p_{(2)}$ with $p_{(1)}\ge p_{(2)}$, and spawn a new expert when a sufficient fraction of files remains persistently difficult.
We cap its initial routing mass (cold-start) and use a higher learning rate for the new expert and a lower learning rate for old experts to accelerate adaptation without forgetting.

\subsection{Motion Generation}
For motion generation, we use filtered text-annotated motion sequences from HumanML3D~\cite{Guo_2022_HUMANML3D} that are retargeted to our robot embodiment for training.
The paired motion–text data provide fine-grained language supervision for training the robot-native diffusion model.
A 262-dimensional per-frame motion descriptor for the robot body is defined as $m_t \in \mathbb{R}^{262}$:
\begin{equation}
m_t = \left[\, \dot{\omega}_t^{\text{root}},\; \dot{v}_t^{\text{root}},\; z_t,\; p_t^{\text{ric}},\; R_t^{\text{6d}},\; \dot{p}_t^{\text{local}},\; c_t^{\text{foot}},\; c_t^{\text{hand}} \,\right].
\end{equation}
Here, $\dot{\omega}_t^{\text{root}} \in \mathbb{R}^{3}$ denotes the root angular velocity, $\dot{v}_t^{\text{root}} \in \mathbb{R}^{3}$ the root linear velocity, and $z_t \in \mathbb{R}$ the root height. The term $p_t^{\text{ric}} \in \mathbb{R}^{36}$ concatenates the local positions of 12 rigid bodies (end-effectors, knees), $R_t^{\text{6d}} \in \mathbb{R}^{174}$ concatenates the 6D rotation representations of 29 rigid bodies (corresponding to the robot's 29 DOF), and $\dot{p}_t^{\text{local}} \in \mathbb{R}^{39}$ concatenates the local velocities of 13 rigid bodies (end-effectors, knees, root). Finally, $c_t^{\text{foot}} \in \{0,1\}^{4}$ and $c_t^{\text{hand}} \in \{0,1\}^{2}$ are binary contact indicators for feet and hands, respectively. All features are extracted under a canonical heading (the first frame faces $+X$). Since the elements of $R_t^{\text{6d}}$ naturally lie in $[-1,1]$, no additional normalization is applied in order to preserve the geometric semantics of rotations.

A conditional denoising diffusion framework is adopted. DistilBERT~\cite{koroteev2021bert} encodes the text into token embeddings $\{c_i\}_{i=1}^{N}$, and the denoising network is implemented as an AdaLN Transformer Decoder, where the diffusion timestep is injected via AdaLN. The training objective is
\begin{equation}
\mathcal{L}_{\text{diff}} = \mathbb{E}_{m_0, t, \epsilon}\left[\left\| m_0 - \hat{m}_\theta(m_t, t, l) \right\|^2\right],
\end{equation}
where $t$ denotes the diffusion timestep (distinct from the frame index in $m_t$), $\epsilon$ is the injected noise, and $l$ is the conditioning signal derived from the text.

\textbf{Token-level Parameter-mixing Mixture of Experts (TP-MoE).}
TP-MoE is inserted after the FFN of each decoder layer to enable fine-grained alignment between individual text tokens and the motion timeline. An expert pool $\{e_1, \ldots, e_K\}$ is maintained. For each text token embedding $c_i$, a gating network produces expert weights, and experts are mixed in parameter space:
\begin{equation}
\omega_i = \text{softmax}(\mathcal{G}(c_i)), \quad \hat{e}^{(i)} = \sum_{k=1}^{K} \omega_{i,k} \cdot e_k,
\end{equation}
where $\omega_i \in \mathbb{R}^K$ denotes the expert weights for token $c_i$, $e_k$ is the $k$-th expert (a two-layer FFN), and $\hat{e}^{(i)}$ is the resulting token-specific mixed expert.

After transforming the motion features, a spatial mask is applied based on the cross-attention weights $A$:
\begin{equation}
M_{t,i} = \sigma\left(\gamma(A_{t,i} - \beta \cdot \max_{t'}A_{t',i})\right),\;\; \Delta x = \sum_{i} M_i \odot \hat{e}^{(i)}(x),
\end{equation}
where $A_{t,i}$ is the attention weight from token $i$ to frame $t$, $\gamma$ and $\beta$ control mask sharpness and thresholding, $\sigma(\cdot)$ is the sigmoid function, and $\odot$ denotes element-wise multiplication. For notational convenience, $M_i$ stacks $\{M_{t,i}\}_t$ along the temporal dimension. The resulting update $\Delta x$ is injected into the backbone through a residual connection. During training, a load-balancing loss $\mathcal{L}_{\text{bal}}$ is introduced to prevent expert collapse.

\textbf{Action-Semantic Frequency-aware Oversampling (ASFO).}
To address the long-tailed distribution of motion data, ASFO leverages an LLM to extract a set of action-semantic tags $\mathcal{K}$ from the text annotations. For each tag, its empirical frequency is computed as $f_m$, and the median frequency $\tau=\mathrm{median}(\{f_m\})$ is used as the target to derive an oversampling multiplier:
\begin{equation}
\rho_m=\min\!\left(\left\lfloor \tau/f_m \right\rceil,\rho_{\max}\right),
\end{equation}
where $\rho_{\max}$ caps the multiplier to limit overfitting. For a multi-label sample $x_j$, the effective multiplier is set to
\begin{equation}
r_j=\max_{k_m\in\phi(x_j)} \rho_m,
\end{equation}
where $\phi(x_j)$ returns the tag set associated with $x_j$. To further enhance diversity for scarce semantics without inflating already frequent actions, we augment data via left--right mirroring \emph{only} for rare-tag samples: for each draw of $x_j$, we mirror joint channels by swapping left/right counterparts (and contact signals when applicable) and use the mirrored sample with a rarity-dependent probability increasing with $r_j$ (e.g., $p_{\mathrm{mir}}(x_j)=\min(\alpha(r_j-1),1)$), while keeping the original otherwise; if side-specific tags are present, we swap left/right tags accordingly. Overall, this strategy ensures that rare actions receive strong and diverse training signals.

\newcommand{\std}[1]{\textsuperscript{\tiny$\pm$#1}} 
\newcommand{\rt}{$^{\dagger}$} 

\begin{table*}[t]
\centering
\small
\setlength{\tabcolsep}{4pt}
\renewcommand{\arraystretch}{1.15}
\caption{
Comparison of motion generation methods under the retarget setting.
$^{\dagger}$ denotes evaluation under the retarget setting.
$\uparrow/\downarrow$ indicate higher/lower is better; $\rightarrow$ denotes a reference metric
\textbf{Bold} and \underline{underlined} values denote the best and second-best results, respectively.
Results are reported as mean $\pm$ standard deviation over five generator rollout seeds.
}
\label{tab:motion_gen}

\begin{tabular}{lccccccc}
\toprule
Motion Generation 
& FID$\downarrow$ 
& R@3$\uparrow$ 
& MM-Dist$\downarrow$ 
& Diversity$\uparrow$ 
& Penetration (mm)$\downarrow$ 
& Floating (mm)$\rightarrow$ 
& Skating$\downarrow$ \\
\midrule

GT\rt(HumanML)~\cite{Guo_2022_HUMANML3D}
& 0.064\std{0.0058} 
& 0.7812\std{0.0207} 
& 1.144\std{0.0105} 
& 1.406\std{0.0141} 
& 0.03 
& 23.64 
& 5.15\% \\
\midrule

T2M-GPT\rt~\cite{zhang2023t2mgpt}
& 0.3782\std{0.0094} 
& 0.5234\std{0.0070} 
& 1.3859\std{0.0024} 
& 1.2501\std{0.0086} 
& 1.14
& 163.80
& 2.77\% \\

MLD\rt~\cite{chen2023executing}
& 0.4060\std{0.0092} 
& 0.4962\std{0.0055} 
& 1.4067\std{0.0082} 
& 1.2054\std{0.0076} 
& 25.11 
& 0.00 
& 34.87\% \\

MDM\rt~\cite{tevet2022human}
& \underline{0.2550}\std{0.0065} 
& 0.6156\std{0.0172} 
& \textbf{1.3143}\std{0.0037} 
& \textbf{1.3355}\std{0.0049} 
& 5.12 
& 64.49 
& 19.16\% \\

MotionGPT\rt~\cite{jiang2023motiongpt}
& 0.2963\std{0.0089} 
& 0.6023\std{0.0023} 
& 1.3651\std{0.0050} 
& 0.8538\std{0.0023} 
& 1.49 
& 16.28 
& 9.49\% \\

Closd\rt~\cite{tevet2024closd}
& 0.3165\std{0.0066} 
& \textbf{0.6208}\std{0.0124} 
& 1.3329\std{0.0029} 
& \underline{1.2580}\std{0.0069} 
& 3.41 
& 45.32
& 20.01\% \\

Closd-Physics\rt
& 0.3740\std{0.0104} 
& 0.5356\std{0.0065} 
& 1.4066\std{0.0017} 
& 1.1503\std{0.0056} 
& \underline{0.82}
& 15.28 
& \textbf{1.21\%} \\

\midrule

TextOp~\cite{xie2026textop}
& 0.3074\std{0.0011} 
& 0.4975\std{0.0039} 
& 1.4009\std{0.0013} 
& 0.7467\std{0.0022} 
& \textbf{0.00} 
& 26.28
& 7.5\% \\

Ours w/o TP-MOE 
& 0.2297\std{0.0069} 
& 0.5276\std{0.0067} 
& 1.3857\std{0.0087} 
& 1.0522\std{0.0124} 
& 2.42 
& 24.36
& 8.7\% \\

Ours 
& \textbf{0.1823}\std{0.0082} 
& \underline{0.6176}\std{0.0063} 
& \underline{1.3302}\std{0.0033} 
& 1.1147\std{0.0082} 
& 3.24
& 32.43
& 8.2\%\\

Ours [Fine-tuned]
& 0.2017\std{0.0021} 
& 0.5702\std{0.0055} 
& 1.3659\std{0.0041} 
& 1.1047\std{0.0023} 
& \textbf{0.00}
& 19.39
& \underline{1.58\%}\\

\bottomrule
\end{tabular}

\end{table*}

\subsection{Physics-Prefix-Guided Motion Generation Fine-tuning}

\textbf{Physics-Guided Prefix Conditioning.}
To improve dynamic feasibility during diffusion sampling, we condition the motion generator on a physically executable prefix.
At inference time, a motion segment $x^{\text{prefix}}$ fine-tuned by the GMT simulator is concatenated with the desired terminal constraint $x^{\text{target}}$ (e.g., a standing pose) and provided as conditioning context to the diffusion model:
\begin{equation}
x_{1:T} \sim p_{\theta}\!\left(x_{1:T} \mid x^{\text{prefix}},\, x^{\text{target}}\right).
\end{equation}
Anchoring the denoising process around a dynamically consistent prefix steers sampling toward locally feasible regions of the state space and mitigates unstable rollouts.

\textbf{Closed-Loop Refinement and Diffusion Fine-tuning.}
We validate each generated motion in simulation using our trained GMT tracker and measure the mean per-joint position error (MPJPE), i.e., the average Euclidean distance between tracked and target joint positions over joints and time.
Trajectories whose MPJPE exceeds a tolerance are rejected and resampled, while feasible rollouts are retained.
This generate--simulate--select loop suppresses common failure modes such as penetration, loss of balance, and long-horizon drift by explicitly filtering dynamically inconsistent samples.

To further align the generative distribution with physically executable motion, we perform progressive fine-tuning after our pretrained GMT by iteratively applying PPO updates to the Stage-II GMT policy on prefix-conditioned closed-loop rollouts, while keeping the diffusion generator frozen.
At each iteration, conditioned on the current prefix, we sample a feasible 1-second continuation, and the simulator then performs end-to-end refinement on the concatenated window consisting of the prefix and the newly generated 1-second segment.
We append the refined continuation to the prefix and repeat this receding-horizon procedure until reaching the designated suffix.
By progressively extending the conditioned context under closed-loop filtering, the denoising process remains stable, and the resulting motions stay kinematically plausible and dynamically consistent over long horizons. Finally, we deploy the physically executable motions together with our fine-tuned GMT controller to the physical robot, achieving robust tracking of agile whole-body behaviors.

%% file: sections/4exp.tex
\begin{figure}[t]
  \centering
  \includegraphics[width=0.9\columnwidth]{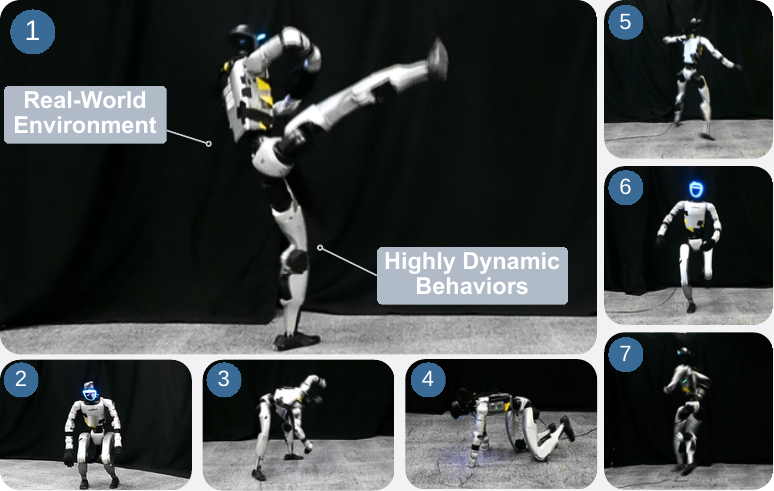}
  \caption{Qualitative results on real robots demonstrating agile, whole-body motion generation across diverse behaviors.}
  \label{fig:real_robot_results}
\end{figure}
\section{Experiments}
We evaluate \textbf{PhyGile} through comprehensive offline and real-robot experiments, demonstrating stable tracking and execution of semantically aligned, high-difficulty agile humanoid motions. Additional qualitative results are provided in Fig.~\ref{fig:real_robot_results} and the accompanying video.
\subsection{Datasets and Evaluation Metrics}
\textbf{Datasets.} The motion generation model is evaluated on a robot-retargeted version of the standard HumanML3D~\cite{Guo_2022_HUMANML3D} test split. The General Motion Tracking (GMT) policy is assessed on a retargeted AMASS~\cite{mahmood2019amass} test set to examine the controller’s ability to track highly dynamic human motions.

\textbf{Evaluation Metrics.} We comprehensively evaluate our framework across both semantic generation quality and physical execution fidelity. The metrics are divided into two categories: Robot Motion Generation and Motion Tracking.

For robot motion generation, we evaluate semantic quality and physical realism.
(1) \textbf{FID} measures feature-distribution fidelity between generated and real motions.
(2) \textbf{R@3} evaluates text--motion alignment via Top-3 retrieval accuracy in the shared feature space.
(3) \textbf{MM-Dist} quantifies cross-modal consistency by the distance between text and motion embeddings.
(4) \textbf{Diversity} is computed as feature-level variance to reflect generation richness.
(5) \textbf{Penetration} reports robot--environment interpenetration depth (mm).
(6) \textbf{Floating} measures unintended airborne height (mm).
(7) \textbf{Skating} measures tangential foot slippage during planted contacts. All generation metrics are trained in the same robot representation space (3 root translations + 4D root orientation + 29 DoF).

For Motion Tracking Evaluation, we quantify tracking accuracy and execution stability.
(1) \textbf{MPJPE} ($E_{mpjpe}$, m) measures mean per-joint position error.
(2) \textbf{MPJAE} ($E_{mpjae}$, rad) measures mean per-joint angle error.
(3) \textbf{MPJVE} ($E_{mpjve}$, rad/s) measures mean per-joint velocity error.
(4) \textbf{Success Rate} is the fraction of episodes without terminal failures: pelvic $Z$ deviation $>0.3$m, trunk gravity-projection difference $>0.8$, or end-effector $Z$ error $>0.3$m.

\subsection{Offline Evaluation}

\subsubsection{Motion Generation Evaluation}
We compare PhyGile against state-of-the-art baselines across two paradigms: human motion generation models that map generated motions to the robot via General Motion Retargeting ~\cite{joao2025gmr}, and robot motion generation approaches (e.g., TextOp~\cite{xie2026textop}) that synthesize motion directly within the robot's kinematic space.

Table~\ref{tab:motion_gen} shows a clear semantics--physics trade-off under retargeting. Human-motion generators inherit strong alignment and diversity from large human datasets, but often produce severe physical artifacts (penetration/floating/skating) after retargeting due to embodiment mismatch. Robot-space methods like TextOp and our PhyGile reduce these violations, yet the tighter feasible set induced by kinematic, balance, and contact constraints can hurt text alignment and diversity.

PhyGile resolves this tension by generating directly in robot space with TP-MOE: \emph{Ours} achieves the best distribution matching (FID) while maintaining strong semantic fidelity (R@3, MM-Dist) and plausible physics. With simulator-based physical fine-tuning, \emph{Ours [Fine-tuned]} further improves contact safety (penetration/skating) via simulator fine-tuning, at a minor cost to retrieval and distribution metrics. Removing TP-MOE (\emph{Ours w/o TP-MOE}) degrades alignment and distribution matching, and Fig.~\ref{fig: ablation}(a) shows moderate expert activation in our TP-MOE is optimal (R@3 saturates around top-$k{=}6$).

\begin{table}[t]
\centering
\small
\setlength{\tabcolsep}{6pt}
\caption{Quantitative evaluation on General Motion Tracking. Each component of our two-stage curriculum contributes to robust motion tracking, and the full PhyGile pipeline achieves the most stable and reliable execution.}
\label{tab: expgmt}
\begin{tabular}{lcccc}
\toprule
Method & $E_{mpjpe}\downarrow$ & $E_{mpjae}\downarrow$ & $E_{mpjve}\downarrow$ & Success$\uparrow$ \\
\midrule
GMT~\cite{chen2025general} & 0.6711 & 0.1098 & 0.6080 & \underline{0.8914} \\
TextOp~\cite{xie2026textop} & \textbf{0.2427} & 0.0927 & \underline{0.4824} & 0.8888 \\
\midrule
PhyGile-C & 0.4960 & 0.0910 & 0.4948 & 0.8537 \\
PhyGile-CF & 0.4322 & 0.0920 & 0.4995 & 0.8629 \\
PhyGile-CFM & 0.4523 & \underline{0.0873} & 0.5014 & 0.8826 \\
\textbf{PhyGile} & \underline{0.2566} & \textbf{0.0720} & \textbf{0.4222} & \textbf{0.9401} \\
\bottomrule
\end{tabular}
\end{table}

\subsubsection{Motion Tracking Evaluation}
We evaluate the robustness of our motion tracking controller against the GMT~\cite{chen2025general} and TextOp~\cite{xie2026textop} baselines, and conduct a systematic ablation study to examine the effect of each component in our two-stage training curriculum.

As summarized in Table~\ref{tab: expgmt}, the complete pipeline (\textbf{PhyGile}) achieves the best overall robustness on the test set, attaining the highest success rate and improved stability over the standard GMT~\cite{chen2025general} baseline. Moreover, \textbf{PhyGile} attains the lowest angular and velocity errors among all compared methods. Although TextOp~\cite{xie2026textop} achieves a lower position error, the two methods exhibit different optimization profiles, with \textbf{PhyGile} showing stronger robustness and more consistent performance across metrics.

\begin{figure}[t]
  \centering
  \includegraphics[width=\columnwidth]{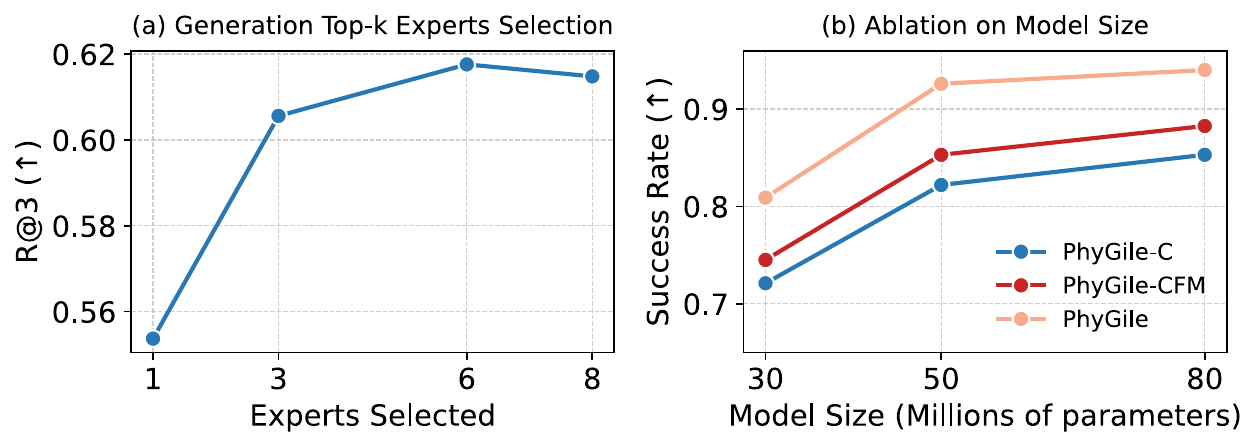}
  \caption{Ablation on key design choices. (a) Generation module: varying the top-k selected experts improves performance up to k=6 (peak R@3), with a slight drop at k=8. (b) GMT module: increasing module size consistently raises the success rate; the full PhyGile outperforms PhyGile-C and PhyGile-CFM across all sizes.}
  \label{fig: ablation}
\end{figure}

To isolate the contribution of each module, we introduce several variants. \textbf{PhyGile-C} employs only difficulty-stratified curriculum learning, yielding a clear improvement over GMT in both tracking accuracy. \textbf{PhyGile-CF} further incorporates the freeze-and-drop self-purification mechanism to filter physically incompatible reference motion, leading to additional gains in stability and overall performance. \textbf{PhyGile-CFM} introduces a Mixture-of-Experts (MoE) design during Stage~I, enabling specialization across difficulty tiers and improving the success rate relative to the curriculum-only setting. Finally, \textbf{PhyGile} applies global soft-MoE fine-tuning on unlabeled data in Stage~II, consistently strengthening robustness and reducing joint errors. Moreover, scaling up the GMT module further boosts the success rate for all variants, while the full \textbf{PhyGile} remains the best across all sizes (Fig.~\ref{fig: ablation}(b)).

\subsection{Real-world Deployment}
\label{sec:deploy}
We deploy our system on a Unitree G1 humanoid robot with 29 degrees of freedom. The fine-tuned general motion tracking policy runs at 50~Hz using ONNX Runtime, taking proprioceptive states together with the current reference motion to output joint-level commands in real time. For deployment, we use the physics-prefix-guided, fine-tuned GMT controller to track a diverse set of agile, high-difficulty whole-body motions on hardware. All trajectories are first verified to be physically executable via sim-to-sim validation and are then streamed as references to the controller for real-time tracking. Notably, a single fine-tuned GMT policy generalizes across a wide range of complex behaviors without per-skill retraining, enabling stable and robust execution.

%% file: sections/5conclusion.tex
\section{Conclusion}
In this paper, we introduced \textbf{PhyGile}, a physics-grounded framework for text-conditioned agile motion generation and execution on humanoid robots. The proposed framework couples a language-driven diffusion generator in robot-skeleton space with a general motion tracker, using physics-validated motion prefixes as a principled interface between them. By anchoring sampling with executable prefixes and fine-tuning the tracking policy with the generated motion clips, it narrows the gap between semantic intent and dynamic feasibility, yielding robust performance on challenging agile behaviors. Comprehensive evaluations in simulation and on hardware show that \textbf{PhyGile} achieves stronger text–motion consistency and more stable, higher-success execution than prior pipelines, establishing an effective route from open-vocabulary language commands to physically realizable whole-body skills.

%% file: sections/appendix.tex
\newpage
\appendix
\subsection{\large\bfseries Motion Generation Module}

The motion generation module is a diffusion-based generator that takes text prompts as input and outputs 262-dimensional robot-native motion sequences. We adopt a Transformer decoder architecture with AdaLN for timestep injection and TP-MoE for fine-grained text-motion alignment. The model is trained on retargeted HumanML3D data with ASFO to handle long-tailed action distributions.

\vspace{2px}
{\normalsize\bfseries 1. Network Architecture.}

The denoising network is an 8-layer AdaLN Transformer decoder with latent dimension $d=512$, feed-forward dimension 1024, 4 attention heads, and dropout rate 0.1. We use GELU activation throughout.

\paragraph{Text Encoder and Summary Token}
We use frozen DistilBERT~\cite{koroteev2021bert} as the text encoder, outputting variable-length 768-dim token embeddings. A learnable query vector performs multi-head attention pooling (4 heads) over these tokens to produce a single summary embedding, which is concatenated with the original tokens to form the cross-attention memory of shape $[(1{+}N)\times d]$. This provides both a global semantic signal and fine-grained per-token information.

\paragraph{TP-MoE Configuration}
We use $K=12$ experts, each being a two-layer FFN with hidden dimension 1024. The gating network $\mathcal{G}$ is a 3-layer MLP ($768 \to d \to d \to K$, with SiLU activations) that maps DistilBERT token embeddings to expert weights. The spatial mask parameters are $\gamma=24$ (sharpness) and $\beta=0.25$ (threshold ratio). The load-balancing loss weight is $\lambda_{\text{lb}}=0.01$, computed as $\mathcal{L}_{\text{lb}} = K \sum_{j=1}^{K}(\bar{p}_j - 1/K)^2$, where $\bar{p}_j$ is the mean routing probability for expert $j$.

\vspace{2px}
{\normalsize\bfseries 2. Feature Representation Details.}
\setcounter{paragraph}{0}

\paragraph{Body Selection}
For the 262-dimensional representation, we select 12 informative bodies for $p_t^{\text{ric}}$: left/right elbow, wrist\_roll, knee, ankle\_pitch, ankle\_roll, and palm (body indices $[7,8,12,13,17,18,19,23,24,25,28,29]$ in the 30-body skeleton). Similarly, $\dot{p}_t^{\text{local}}$ includes these 12 bodies plus the root (13 bodies total). This selection excludes nearly-static torso bodies (chest, abdomen, shoulders, hips) while retaining all articulations and end-effectors.

\paragraph{Block-wise Normalization}
Z-score normalization is applied block-wise: $\dot{\omega}_t^{\text{root}}$ (dims 0--2), $\dot{v}_t^{\text{root}}$ (dims 3--5), $z_t$ (dim 6), $p_t^{\text{ric}}$ (dims 7--42), and $\dot{p}_t^{\text{local}}$ (dims 217--255) are normalized; $R_t^{\text{6d}}$ (dims 43--216) is \emph{not} normalized since its elements naturally lie in $[-1,1]$; binary contact indicators $c_t^{\text{foot}}$ (dims 256--259) and $c_t^{\text{hand}}$ (dims 260--261) are preserved without normalization.

\paragraph{Contact Detection}
$c_t^{\text{foot}} \in \{0,1\}^4$ (4 values for left/right ankle pitch/roll) is set to 1 when ankle height $< 0.05$\,m and horizontal velocity $< 0.01$\,m/s. $c_t^{\text{hand}} \in \{0,1\}^2$ uses a height threshold of $0.10$\,m.

\paragraph{Canonical Heading}
All sequences are rotated so the first frame faces $+X$ (yaw $=0$). Only the yaw component is removed; pitch and roll are preserved, which is critical for non-upright motions (e.g., rolling, crawling). The root xy position is also shifted to the origin at the first frame.

\vspace{2px}
{\normalsize\bfseries 3. Classifier-Free Guidance with Negative Prompts.}

We support both standard CFG and negative-prompt CFG. In standard mode:
\begin{equation}
\hat{m}_0 = \hat{m}_0^{\varnothing} + s \cdot (\hat{m}_0^{l} - \hat{m}_0^{\varnothing}),
\end{equation}
where $\hat{m}_0^{\varnothing}$ is the unconditional prediction (text embedding zeroed out via the 10\% training dropout) and $s=2.5$. When a negative prompt $l^{-}$ is provided, its encoding replaces the unconditional output:
\begin{equation}
\hat{m}_0 = \hat{m}_0^{l^{-}} + s \cdot (\hat{m}_0^{l} - \hat{m}_0^{l^{-}}),
\end{equation}
which steers generation \emph{away} from the negative description (e.g., ``no walking, no standing still'') while following the positive prompt.

\subsection{\large\bfseries General Motion Tracking}

This section provides additional implementation details of the General Motion Tracking (GMT) module, including the complete reward specification, observation and action spaces, future command encoder, adaptive sampling strategy, curriculum hyperparameters, domain randomization, early termination conditions, PPO settings, network architecture, simulation setup, and the neuro-control difficulty taxonomy.

\vspace{2px}
{\normalsize\bfseries 1. Reward Functions.}
\setcounter{paragraph}{0}

The reward function used for GMT training consists of task rewards for accurate motion imitation and regularization rewards for smooth and safe control. All task reward terms adopt an exponential kernel of the form
\begin{equation}
r = \exp\left(-\frac{e}{\sigma^2}\right),
\end{equation}
where $e$ denotes the squared error term and $\sigma$ controls the reward sensitivity.

\paragraph{Task rewards}
The task reward terms are summarized in Table~\ref{tab:gmt_task_rewards}. Here, $\mathbf{p}$ and $\hat{\mathbf{p}}$ denote the reference and simulated positions, respectively; $\mathbf{q} \ominus \hat{\mathbf{q}}$ denotes the quaternion error magnitude; $\mathbf{v}$ and $\boldsymbol{\omega}$ denote linear and angular velocities; and $N$ is the number of tracked rigid bodies.

\begin{table}[H]
\centering
\caption{Task reward terms used in GMT training.}
\label{tab:gmt_task_rewards}
\resizebox{\linewidth}{!}{
\begin{tabular}{llll}
\toprule
Term & Expression & Weight & $\sigma$ \\
\midrule
Global Anchor Position &
$\exp\left(-\frac{\|\mathbf{p}^{\text{anchor}} - \hat{\mathbf{p}}^{\text{anchor}}\|^2}{\sigma^2}\right)$
& 0.8 & 0.2 \\

Global Anchor Orientation &
$\exp\left(-\frac{\|\mathbf{q}^{\text{anchor}} \ominus \hat{\mathbf{q}}^{\text{anchor}}\|^2}{\sigma^2}\right)$
& 0.5 & 0.4 \\

Relative Body Position &
$\exp\left(-\frac{\frac{1}{N}\sum_{i}\|\mathbf{p}_i - \hat{\mathbf{p}}_i\|^2}{\sigma^2}\right)$
& 1.0 & 0.3 \\

Relative Body Orientation &
$\exp\left(-\frac{\frac{1}{N}\sum_{i}\|\mathbf{q}_i \ominus \hat{\mathbf{q}}_i\|^2}{\sigma^2}\right)$
& 1.0 & 0.4 \\

Body Linear Velocity &
$\exp\left(-\frac{\frac{1}{N}\sum_{i}\|\mathbf{v}_i - \hat{\mathbf{v}}_i\|^2}{\sigma^2}\right)$
& 1.0 & 1.0 \\

Body Angular Velocity &
$\exp\left(-\frac{\frac{1}{N}\sum_{i}\|\boldsymbol{\omega}_i - \hat{\boldsymbol{\omega}}_i\|^2}{\sigma^2}\right)$
& 1.0 & 3.14 \\
\bottomrule
\end{tabular}
}
\end{table}

The \emph{Global Anchor Position} and \emph{Global Anchor Orientation} terms measure pelvis tracking error in world coordinates. The \emph{Relative Body Position} and \emph{Relative Body Orientation} terms are computed relative to the anchor frame and therefore capture body-pose consistency independently of global placement. The \emph{Body Linear Velocity} and \emph{Body Angular Velocity} terms encourage temporal consistency with the reference motion.

\paragraph{Regularization rewards}
The regularization terms are listed in Table~\ref{tab:gmt_reg_rewards}.

\begin{table}[H]
\centering
\caption{Regularization rewards used in GMT training.}
\label{tab:gmt_reg_rewards}
\begin{tabular}{lll}
\toprule
Term & Expression & Weight \\
\midrule
Action Rate L2 & $\|a_t - a_{t-1}\|_2^2$ & $-0.1$ \\
Joint Limit Penalty & $\sum_{j} \mathbf{1}_{\text{out}}(q_j)$ & $-10.0$ \\
Undesired Contacts & $\sum_{b} \mathbf{1}_{F_b > 1.0}$ & $-0.1$ \\
\bottomrule
\end{tabular}
\end{table}

The \emph{Action Rate L2} term penalizes abrupt changes in joint-position targets to promote smooth control signals. The \emph{Joint Limit Penalty} applies a large penalty whenever any joint exceeds its feasible position range, thereby improving mechanical safety. The \emph{Undesired Contacts} term penalizes contact forces larger than 1.0~N on non-end-effector bodies. The following bodies are excluded from this penalty:
\texttt{left\_ankle\_roll\_link},
\texttt{right\_ankle\_roll\_link},
\texttt{left\_wrist\_yaw\_link}, and
\texttt{right\_wrist\_yaw\_link}.

\vspace{2px}
{\normalsize\bfseries 2. Observation and Action Spaces}
\setcounter{paragraph}{0}

\paragraph{Policy observation space}
The policy observation has 616 dimensions and is summarized in Table~\ref{tab:gmt_policy_obs}. Additive uniform noise is applied during training for domain randomization: $\pm 0.05$ on root orientation, $\pm 0.2$ on angular velocity, $\pm 0.01$ on joint positions, and $\pm 0.5$ on joint velocities.

\begin{table}[H]
\centering
\caption{Policy observation space for our GMT.}
\label{tab:gmt_policy_obs}
\begin{tabular}{l p{3.5cm} c}
\toprule
Component & Description & Dimension \\
\midrule
command & Motion target (current + future frames) & 520 \\
motion\_anchor\_ori\_b & Root orientation error in body frame (6D rotation) & 6 \\
base\_ang\_vel & Base angular velocity & 3 \\
joint\_pos & Joint positions (relative to default) & 29 \\
joint\_vel & Joint velocities & 29 \\
actions & Previous actions & 29 \\
\bottomrule
\end{tabular}
\end{table}

\paragraph{Critic observation space}
The critic receives privileged, noise-free observations, as summarized in Table~\ref{tab:gmt_critic_obs}.

\begin{table}[H]
\centering
\caption{Critic observation space for our GMT.}
\label{tab:gmt_critic_obs}
\resizebox{\linewidth}{!}{
\begin{tabular}{l p{3.5cm} c}
\toprule
Component & Description & Dimension \\
\midrule
command & Motion target (current + future frames) & 520 \\
motion\_anchor\_pos\_b & Root position error in body frame & 3 \\
motion\_anchor\_ori\_b & Root orientation error in body frame (6D rotation) & 6 \\
body\_pos & Key body positions ($14$ bodies $\times 3$) & 42 \\
body\_ori & Key body orientations ($14$ bodies $\times 6$D) & 84 \\
base\_lin\_vel & Base linear velocity & 3 \\
base\_ang\_vel & Base angular velocity & 3 \\
joint\_pos & Joint positions & 29 \\
joint\_vel & Joint velocities & 29 \\
actions & Previous actions & 29 \\
\bottomrule
\end{tabular}
}
\end{table}

\paragraph{Motion command structure}
The motion command has 520 dimensions. Each frame contains 65 dimensions, including joint positions (29), joint velocities (29), root position (3), and root quaternion (4). The command structure is summarized in Table~\ref{tab:gmt_command}.

\begin{table}[H]
\centering
\caption{Structure of the motion command used in GMT.}
\label{tab:gmt_command}
\resizebox{\linewidth}{!}{
\begin{tabular}{llll}
\toprule
Segment & Frames & Per-Frame Dim & Total Dim  \\
\midrule
Current & 1 & 65 & 65 \\
Short-horizon & 2 & 65 & 130\\
Long-horizon & 5 & 65 & 325 \\
\bottomrule
\end{tabular}
}
\end{table}

The short-horizon frames are directly concatenated into the observation. The long-horizon frames are sampled with a stride $k=20$, corresponding to $5 \times 20 = 100$ simulation steps, i.e., approximately 2 seconds at 50~Hz, and are processed by the future command encoder.

\paragraph{Action space}
The action is a 29-dimensional vector of joint position targets. A PD controller converts these targets into joint torques at each simulation step. The control frequency is 50~Hz, corresponding to a simulation time step of 0.005~s and a control decimation factor of 4.

\vspace{2px}
{\normalsize\bfseries 3. Error-Driven Adaptive Sampling}
\setcounter{paragraph}{0}

In addition to curriculum-level scheduling, GMT employs file-level adaptive sampling within each difficulty level. For each motion file $i$, the following statistics are maintained using exponential moving averages (EMA):
\begin{align}
E_i &\leftarrow (1-\alpha) E_i + \alpha \tilde{e}_i, \\
\hat{p}_i^{\text{succ}} &= \frac{S_i}{S_i + F_i + \epsilon},
\end{align}
where $\tilde{e}_i$ is the mean tracking error for file $i$ in the current batch, $\alpha = 0.25$, and the success-rate EMA uses decay $\beta = 0.4$.

The sampling score is defined as
\begin{equation}
r_i = (1 - w) \cdot \min(E_i / c, 1) + w \cdot (1 - \hat{p}_i^{\text{succ}}),
\end{equation}
where $c = 0.3$ is the error normalization constant and $w = 0.15$ is the success-rate mixing weight, activated after a 6000-iteration warmup. The final sampling distribution is given by
\begin{equation}
p_i = (1-\varepsilon)\cdot \text{softmax}\left(\frac{\log(r_i+\epsilon)}{T}\right) + \frac{\varepsilon}{N},
\end{equation}
where $T = 1.05$ and $\varepsilon = 0.20$ ensures minimum coverage.

\vspace{2px}
{\normalsize\bfseries 4. Curriculum Learning Hyperparameters}
\setcounter{paragraph}{0}

\paragraph{Metric convergence detection}
Early promotion is enabled when the relative improvement in MPJPE/MPJAE falls below 3\% for 3 consecutive evaluations, subject to a minimum of 3000 iterations on the current level.

\paragraph{Gradual file introduction}
When a new level is unlocked, files are introduced progressively according to
\begin{equation}
r(t) = r_{\text{start}} + \frac{\min(t - t_{\text{unlock}},\, T_{\text{intro}})}{T_{\text{intro}}} \cdot (1 - r_{\text{start}}),
\end{equation}
where $r_{\text{start}} = 0.2$. The associated hyperparameters are listed in Table~\ref{tab:gmt_curriculum_intro}.

\begin{table}[H]
\centering
\caption{Gradual file introduction hyperparameters.}
\label{tab:gmt_curriculum_intro}
\begin{tabular}{lc}
\toprule
Parameter & Value \\
\midrule
Start ratio $r_{\text{start}}$ & 0.2 \\
Base introduction iterations $T_{\text{intro}}$ & 3000 \\
Extra introduction iterations (level $\geq 4$) & +2000 \\
\bottomrule
\end{tabular}
\end{table}

\paragraph{Freeze-and-drop parameters}
The freeze-and-drop mechanism is configured as shown in Table~\ref{tab:gmt_freeze_drop}.

\begin{table}[H]
\centering
\caption{Freeze-and-drop hyperparameters.}
\label{tab:gmt_freeze_drop}
\begin{tabular}{lc}
\toprule
Parameter & Value \\
\midrule
Error threshold $\tau_{\text{err}}$ & 0.1 \\
Success threshold $\tau_{\text{succ}}$ & 0.15 \\
Minimum attempts $n_{\min}$ & 20000 \\
Freeze duration (iterations) & 4000 \\
Freezes before permanent drop & 2 \\
Check interval & 500 iterations \\
\bottomrule
\end{tabular}
\end{table}

\vspace{2px}
{\normalsize\bfseries 5. Domain Randomization}
\setcounter{paragraph}{0}

The domain randomization settings are summarized in Table~\ref{tab:gmt_domain_randomization}.

\begin{table}[H]
\centering
\caption{Domain randomization settings used in GMT training.}
\label{tab:gmt_domain_randomization}
\resizebox{\linewidth}{!}{
\begin{tabular}{lll}
\toprule
Randomization & Range / Parameters & Mode \\
\midrule
Ground friction (static) & $[0.3, 1.6]$ & Startup \\
Ground friction (dynamic) & $[0.3, 1.2]$ & Startup \\
Restitution & $[0.0, 0.5]$ & Startup \\
Joint default position offset & $[-0.01, 0.01]$ rad & Startup \\
Torso CoM offset (x) & $[-0.025, 0.025]$ m & Startup \\
Torso CoM offset (y) & $[-0.05, 0.05]$ m & Startup \\
Torso CoM offset (z) & $[-0.05, 0.05]$ m & Startup \\
External push (interval 1--3 s) & lin: $\pm 0.5$ m/s, ang: $\pm 0.78$ rad/s & Interval \\
\bottomrule
\end{tabular}
}
\end{table}

Additionally, observation noise is injected during training, as described in the policy observation space above.

\vspace{2px}
{\normalsize\bfseries 6. PPO Training Hyperparameters}
\setcounter{paragraph}{0}

The PPO hyperparameters used for GMT training are listed in Table~\ref{tab:gmt_ppo}.

\begin{table}[H]
\centering
\caption{PPO training hyperparameters for GMT.}
\label{tab:gmt_ppo}
\begin{tabular}{lc}
\toprule
Parameter & Value \\
\midrule
Parallel environments & 8192 \\
Steps per env per iteration & 24 \\
Learning epochs per iteration & 5 \\
Mini-batches per epoch & 4 \\
Initial learning rate & $1 \times 10^{-3}$ \\
LR schedule & Adaptive (KL-based) \\
Desired KL divergence & 0.01 \\
Discount factor $\gamma$ & 0.99 \\
GAE $\lambda$ & 0.95 \\
Clip parameter & 0.2 \\
Entropy coefficient & 0.005 \\
Value loss coefficient & 1.0 \\
Max gradient norm & 1.0 \\
Initial action noise std & 1.0 \\
Advantage normalization & Per mini-batch: off \\
Empirical observation normalization & On \\
Maximum iterations & 200{,}000 \\
Checkpoint save interval & 1000 iterations \\
\bottomrule
\end{tabular}
\end{table}

\vspace{2px}
{\normalsize\bfseries 7. Network Architecture}
\setcounter{paragraph}{0}

\paragraph{Actor (MoE policy)}
Each expert MLP uses hidden dimensions $[2048, 1024, 512]$ with ELU activations. The gate network maps the 128-dimensional latent $\mathbf{z}_t$ to $K$ routing logits. During Stage I, the number of active experts grows from 1 to 10 as curriculum levels are unlocked. Stage II may dynamically add experts for persistently difficult motions.

\paragraph{Critic}
The critic is a single MLP with hidden dimensions $[2048, 1024, 512]$ and ELU activations, sharing the same future command encoder as the actor.

\vspace{2px}
{\normalsize\bfseries 8. Simulation Configuration}
\setcounter{paragraph}{0}

The simulation settings are summarized in Table~\ref{tab:gmt_simulation}.

\begin{table}[H]
\centering
\caption{Simulation configuration for GMT training.}
\label{tab:gmt_simulation}
\begin{tabular}{lc}
\toprule
Parameter & Value \\
\midrule
Simulator & Isaac Lab (IsaacSim) \\
Physics dt & 0.005 s \\
Control decimation & 4 (control at 50 Hz) \\
Episode length & 10 s \\
Environment spacing & 2.5 m \\
\bottomrule
\end{tabular}
\end{table}

\begin{table*}[t]
\centering
\caption{Neuro-control difficulty taxonomy used in GMT curriculum design.}
\label{tab:gmt_taxonomy}
\small
\setlength{\tabcolsep}{5pt}
\renewcommand{\arraystretch}{1.15}
\begin{tabular}{p{1.8cm} p{3.6cm} p{4.2cm} p{7.2cm}}
\toprule
Level & Primary Sub-dimensions & Representative Actions & Characteristics \\
\midrule
1 & Low planning & Stand, raise arms, wave hands & Near-static or upper-body only; minimal balance demand \\
2--4 & Planning & Walk, step, turn & Basic locomotion with stable rhythm and simple motor patterns \\
5--7 & Planning + Predictive + Balance & Run, jump, walk backwards & Combined locomotion and dynamic actions requiring higher coordination \\
8 & Planning + Balance & Crawl, kneel, crawl on knees & Low-posture ground movements with increased body control \\
9--10 & All four (maximal) & Cartwheel, backflip, spinning, dance & Acrobatic or high-skill actions with tightly coupled planning, prediction, and balance \\
11 (excluded) & Contextual-dominant & Stair climbing, stepping over obstacles & Requires external terrain or elevation changes \\
12 (excluded) & Contextual-dominant & Swimming, flying & Physically unrealizable in flat-ground simulation \\
\bottomrule
\end{tabular}
\end{table*}

\vspace{2px}
{\normalsize\bfseries 9. Neuro-Control Difficulty Taxonomy}
\setcounter{paragraph}{0}

As discussed in the main text, motion difficulty is interpreted from a neuro-control perspective as the coordination load imposed on the central nervous system (CNS). We decompose this aggregate load into four interpretable sub-dimensions: motor planning load, predictive control load, balance and multisensory integration load, and contextual/environmental constraint load.

\paragraph{Motor Planning Load}
This dimension reflects the complexity of an action's temporal structure and sequential organization. Executing multi-step movement sequences relies on prospective planning and ordinal encoding in the primary motor cortex, premotor cortex, and supplementary motor area (SMA). Actions that require chaining multiple distinct motor primitives in a specific order impose higher planning demands than repetitive or single-phase motions.

\paragraph{Predictive Control Load}
This dimension captures the degree to which an action depends on feedforward prediction and error correction under sensory feedback delays or tight temporal constraints. Actions involving ballistic phases, rapid transitions, or precise timing place a greater demand on predictive control mechanisms.

\paragraph{Balance and Multisensory Integration Load}
This dimension reflects the demand for integrating vestibular, proprioceptive, and visual information to maintain postural stability and spatial orientation. The load increases substantially during dynamic balance, turning, backward locomotion, and aerial phases.

\paragraph{Contextual and Environmental Constraint Load}
This dimension reflects additional perceptual and control demands introduced by terrain structure, elevation changes, or environmental hazards. Such motions require continuous adjustment of gait parameters based on external context and are therefore not fully representable in a flat-ground simulation setting.

\vspace{2px}
{\normalsize\bfseries 10. Description aggregation and primary-action identification}
\setcounter{paragraph}{0}

Each motion instance may be associated with multiple textual annotations that describe the same underlying action from different perspectives or with different levels of detail. During difficulty assessment, these annotations are jointly interpreted to infer the action being performed and to estimate its overall motion complexity. In practice, we use the DeepSeek model to classify the action described by these annotations and to identify the dominant motion primitive underlying the sequence. If a description includes multiple consecutive movement components, we first identify the \emph{primary action}, namely, the most essential or most difficult motion primitive in the sequence, and use it to determine the base difficulty level. Subsequent score adjustment is then based on the number and continuity of the remaining action components. Actions that explicitly depend on external structures (e.g., stairs, platforms, or narrow elevated surfaces) are assigned difficulty level 11, while actions that are physically infeasible under normal gravity conditions, such as swimming or flying, are assigned difficulty level 12 and treated as special cases.

The corresponding mapping to curriculum levels is summarized in Table~\ref{tab:gmt_taxonomy}.